\documentclass{article}
\usepackage{spconf,amsmath,graphicx}
\usepackage{makecell}
\usepackage{amssymb}
\usepackage{graphics} 
\usepackage{epsfig} 
\usepackage{color}
\usepackage{makecell}
\usepackage{multirow}
\usepackage{color}
\usepackage{upgreek}
\usepackage{subfigure}
\usepackage{algorithm}  
\usepackage{algorithmic}  
\usepackage{booktabs}

\title{Multi-loss-aware Channel Pruning of Deep Networks}
\name{Yiming Hu, Siyang Sun, Jianquan Li, Jiagang Zhu, Xingang Wang, Qingyi Gu}
\address{Research Center of Precision Sensing and Control, Institute of Automation, Chinese Academy of Sciences\\ School of Computer and Control Engineering, University of Chinese Academy of Sciences\\
\{huyiming2016, sunsiyang2015, lijianquan2015, zhujiagang2015, xingang.wang, qingyi.gu\}@ia.ac.cn}

\begin{document}
%
\maketitle
\begin{abstract}
Channel pruning, which seeks to reduce the model size by removing redundant channels, is a popular solution for deep networks compression. Existing
channel pruning methods usually conduct layer-wise channel selection by directly minimizing the reconstruction error of feature maps between the baseline model and the pruned one. However, they ignore the feature and semantic distributions within feature maps and real contribution of channels to the overall performance. In this paper, we propose a new channel pruning method by explicitly using both intermediate outputs of the baseline model and the classification loss of the pruned model to supervise layer-wise channel selection. Particularly, we introduce an additional loss to encode the differences in the feature and semantic distributions within feature maps between the baseline model and the pruned one. By considering the reconstruction error, the additional loss and the classification loss at the same time, our approach can significantly improve the performance of the pruned model. Comprehensive experiments on benchmark datasets demonstrate the effectiveness of the proposed method.

\end{abstract}
\section{Introduction}
\label{sec:intro}

In recent years, deep convolutional neural networks (CNNs) have been widely applied to various computer vision tasks, e.g., image classification, object detection, action recognition, since AlexNet \cite{Krizhevsky2012} won the ImageNet Challenge: ILSVRC 2012~{\cite{Russakovsky2015}}. However, it is very hard to deploy these deep models on resource-constrained devices including mobile robots, unmanned aerial vehicles, smartphones, etc. due to their high storage and computational cost.

In order to make CNNs available on resource-constrained devices, many model compression and acceleration methods have been presented including channel pruning \cite{He2017, Luo2017, jiang2018efficient}, network quantization\cite{WangP2018, Xu2018,Zhou2018} and low-rank approximation \cite{Wang2017,Wang2018}. The method proposed in this paper falls into the category of channel pruning. Different from simply making sparse connections, channel pruning reduces the model size by directly removing redundant channels. In contrast, pruning the entire channel can achieve fast inference without special software or hardware implementation. To prune redundant channels, existing reconstruction-based methods \cite{He2017, Luo2017, jiang2018efficient} usually minimize the reconstruction error of feature maps between the baseline model and the pruned one. DCP \cite{zhuang2018discrimination} seeks to conduct channel selection by introducing additional discrimination-aware losses. These methods suffer from two aspects of limitations. First, directly matching feature maps ignores the feature and semantic distributions within them. Specifically, one should be encouraged to focus more on really important channels and spatial activations of feature maps instead of simply aligning every location. Second, the reconstruction error of a layer or additional losses do not always truly reflect the change of the classification loss, which may lead to mistakenly pruning some important channels. 

In this paper, we address these problems from two aspects. First, the feature and semantic correlation loss is defined, through which the pruned model can preserve the feature and semantic distributions within feature maps of the baseline network. Second, we reconsider real contribution of channels to the overall performance. To this end, the classification loss of the pruned model is used to supervise channel selection. Then, we conduct layer-wise channel pruning by considering the reconstruction error, the additional loss and the classification loss at the same time.

Our main contributions are summarized as follows. First, we present a multi-loss-aware channel pruning method for deep networks compression by the introduction of the feature and semantic correlation loss and the classification loss. Second, extensive experiments on benchmark datasets show that the proposed method is theoretically reasonable and practically effective. On CIFAR-10, our approach outperforms the previous state-of-the-art method by $0.55\%$ in accuracy.

\begin{figure*}[t]
	\centering
	\includegraphics[width=1.90\columnwidth]{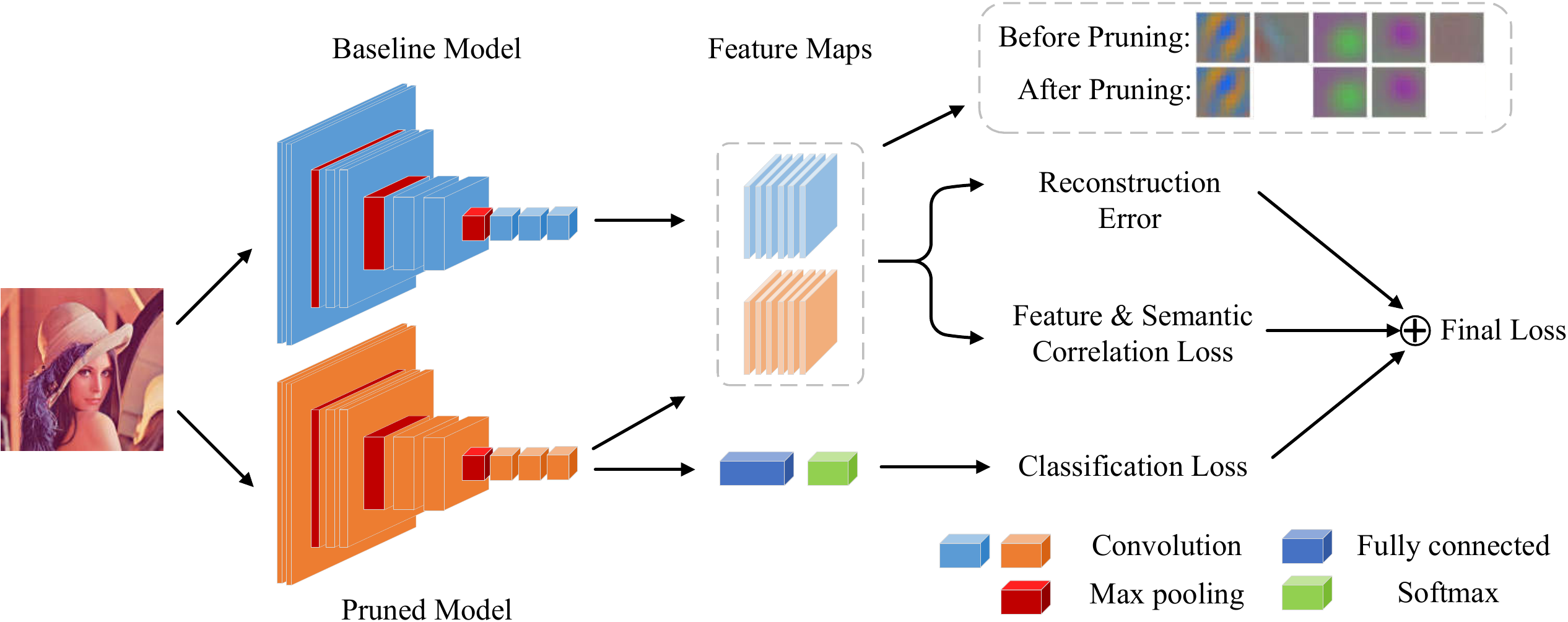}
	\vskip -2mm
	\caption{Illustration of multi-loss-aware channel pruning. The final loss is used to supervise layer-wise channel selection.}
	\label{fig:Main_Framework}
\end{figure*}

\section{Related work}

The basic idea of weight pruning is to prune some redundant weights. Han et al.~{\cite{Han2015}} proposed a conceptually simple pruning method: weights lower than a certain threshold are considered as low contribution ones that can be pruned, and then fine-tuning is taken for restoring the network accuracy. However, the method needs special software or hardware implementation for fast inference. In view of this issue, Structured Sparsity Learning (SSL)~{\cite{Wen2016}} imposed regularizations on different levels of structures such as filters, channels or layers. Then weights belonging to the same level of structure would go to zero at the same time, which could be removed. This method obtains extraordinary performance but is computationally expensive and difficult to converge. Afterwards, some inference-based channel pruning methods \cite{Luo2017, Hu2016} were proposed, and they sought to do channel selection by defining different selection criteria. ThiNet~{\cite{Luo2017}} explicitly formulated channel selection as an optimization problem, and conducted channel pruning by minimizing the reconstruction error of feature maps between the baseline model and pruned one. DCP \cite{zhuang2018discrimination} aimed at selecting the most discriminative channels for each layer by considering both the reconstruction error and the discrimination-aware loss. However, these methods ignore the feature and semantic distributions within feature maps and real contribution of channels to the overall performance.

\section{Our Method}
\label{sec:pagestyle}
For better description of the proposed method, some notations are given first. Considering a layer of the CNN model, $X$ represents the input tensor, $W$ denotes the convolution filter and $\otimes$ represents the convolution operation. We further use $F \in \mathbb{R}^{M\times HZ}$ to denote output feature maps of the layer. Here, $M$, $H$, $Z$ represents the number of output channels, the height and the width of the feature maps respectively.

\subsection{Motivation}
\label{ssec:subhead}

Existing methods \cite{He2017, Luo2017, zhuang2018discrimination} conduct channel pruning by minimizing the reconstruction error of a layer, which can be defined as the Euclidean distance of feature maps between the original model and the pruned one:

\begin{equation}
\mathcal{L}_r = \frac{1}{2T}||X\otimes W-X\otimes W_{\mathcal{P}}||^2_2
\end{equation}
where $T = M \times H \times Z$, $\mathcal{P}$ denotes the index set of the selected channels, $W_{\mathcal{P}}$ represents the submatrix indexed by $\mathcal{P}$. Jiang et al. \cite{jiang2018efficient} explore the correlation between the layer-wise reconstruction error and the classification loss, and they demonstrate these two variables show positive correlation. However, the correlation coefficient varies with different layers and some layers show low correlation. Consequently, simply minimizing the reconstruction error may mistakenly remove some important channels, even if they are very sensitive to the classification loss. Moreover, assuming each channel of $F$ corresponds to a type of feature and each spatial activation encodes a type of semantic information, the feature and semantic distributions reflect which channels and spatial areas a CNN model focuses on. However, directly matching feature maps is not a good choice to preserve the feature and semantic distributions of the baseline model.

These limitations of existing methods inspire us to define a new loss: the feature and semantic correlation loss. The pruned model can preserve the feature and semantic distributions within feature maps of the baseline network using this loss. Meanwhile, we will explicitly use the classification loss to supervise layer-wise channel selection.

\begin{algorithm} 
	\caption{The proposed method}  
	\label{al:channel_pruning}
	\begin{algorithmic}[1] 
		\REQUIRE \{$W^l:0<l<L$\}: parameters of the pre-trained model, the training set \{$x_i,y_i$\}, the pruning rate.
		\ENSURE \{$W^l_{\mathcal{P}}:0<l<L$\}: parameters of the pruned model.
		\STATE Initialize $W^l_{\mathcal{P}}$ with $W^l$ for $\forall 1 \leq l \leq L$
		\FOR{$l = 1, 2,\cdots, L$} 
		\STATE Construct the final loss $\mathcal{L}$ shown as in Fig. \ref{fig:Main_Framework}
		\STATE Perform channel selection for layer $l$ by Eq \ref{eq:sensitivity}
		\STATE Update $W^l_{\mathcal{P}}$ w.r.t. the selected channels by Eq \ref{eq:sgd_step}
		\ENDFOR
		\STATE Fine-tune the pruned model 
	\end{algorithmic} 
\end{algorithm}

\subsection{Formulation}
\label{ssec:subhead}

To transfer rich semantic information of the baseline model, one can align the semantic distribution of two feature maps \cite{huang2017like, zagoruyko2016paying}. In this paper, we consider aligning both the feature and semantic distributions by introducing the feature and semantic correlation loss. Here, the loss can be defined as the square error of two Gram matrices:
\begin{equation}
\mathcal{L}_s = \frac{1}{4N^2M^2}(||G^f - G^f_{\mathcal{P}}||^2_2 + ||G^s - G^s_{\mathcal{P}}||^2_2)
\end{equation}
where $N = H \times Z$, the Gram matrix $G^f\in \mathbb{R}^{M\times M}$ and $G^s\in \mathbb{R}^{N\times N}$ encode the feature and semantic correlation for a layer of the baseline model respectively. Similarly, $G^f_{\mathcal{P}}$ and $G^s_{\mathcal{P}}$ are Gram matrices corresponding to the pruned model. $G^f\in \mathbb{R}^{M\times M}$ is computed by the inner product between different vectorized features. $G^s\in \mathbb{R}^{N\times N}$ is computed by the inner product between different vectorized spatial activations cross channels:
\begin{equation}
G^f_{ij} = \sum_{k=1}^{N}F_{ik} F_{jk}, G^s_{i^{'}j^{'}} = \sum_{k^{'}=1}^{M}F_{k^{'}i^{'}} F_{k^{'}j^{'}}
\end{equation}

To ensure important channels being kept, we explicitly use the classification loss to supervise layer-wise channel selection. By considering the reconstruction error, the feature and semantic correlation loss and the classification loss at the same time, the problem of channel pruning can be formulated to minimize the following joint loss function:

\begin{equation}
\label{eq:FinalLoss}
\begin{split}
\mathop {\min }\limits_{W_{\mathcal{P}}} \quad &\mathcal{L}(W_{\mathcal{P}}) = \mathcal{L}_r(W_{\mathcal{P}}) + \alpha \mathcal{L}_s(W_\mathcal{P}) + \beta \mathcal{L}_c(W_{\mathcal{P}})\\
&s.t. \quad ||\mathcal{P}||_0 \leq \mathcal{K} 
\end{split}
\end{equation}
where $\alpha$ and $\beta$ are two positive coefficients balancing the regularization terms, $\mathcal{L}_c$ is the classification loss and $\mathcal{K}$ is the number of channels to be selected.

\begin{figure}[t]
	\centering
	\includegraphics[width=0.98\columnwidth]{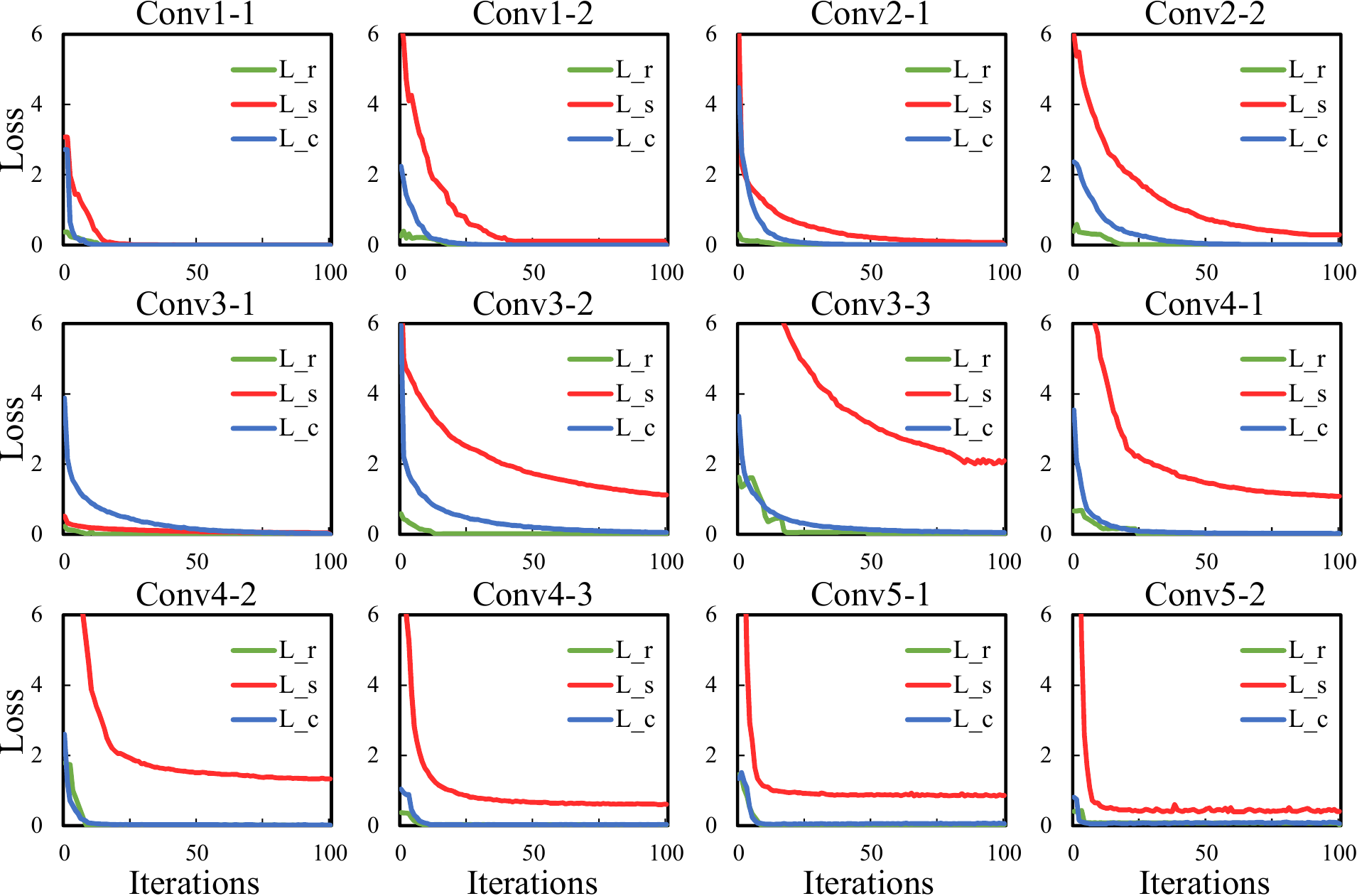}
	\vskip -2mm
	\caption{The evolution curves of different losses over epochs on CIFAR-100.}
	\label{fig:Convergence}
\end{figure}

\subsection{Optimization Algorithm}
\label{ssec:subhead}
It is a combinatorial optimization problem to choose a subset of channels according to Eq. \ref{eq:FinalLoss}. Directly searching competitive network structures in the whole solution space is NP-hard. Existing methods \cite{Luo2017, zhuang2018discrimination, jiang2018efficient} seek to do channel selection with greedy algorithm by only considering the weight magnitude or the gradient of Eq. \ref{eq:FinalLoss} w.r.t. the channel. In this paper, we take both of them into consideration. To be specific, the sensitivity of the $k$-th channel is defined as: 
\begin{equation}
\label{eq:sensitivity}
\delta_k = \sum_{i=1}^{H}\sum_{j=1}^{W}||\frac{\mathcal{L}}{W_{k,i,j}}W_{k,i,j}||^2_2
\end{equation}
which is the square sum of weights of the $k$-th channel multiplying gradients of Eq. \ref{eq:FinalLoss} w.r.t. the channel. Then we retain the channels with the $i$ largest sensitivity and remove others. After this, Eq. \ref{eq:FinalLoss} w.r.t. the selected channels is further optimized according to stochastic gradient descent (SGD). $W_{\mathcal{P}}$ is updated by:

\begin{equation}
\label{eq:sgd_step}
W_{\mathcal{P}} = W_{\mathcal{P}} - \eta \frac{\partial\mathcal{L}}{\partial W_{\mathcal{P}}}
\end{equation}
where $\eta$ represents the learning rate. After updating $W_{\mathcal{P}}$, the channel pruning of a single layer is finished. The pruning process of the whole model is described in Algorithm \ref{al:channel_pruning}.

\renewcommand{\multirowsetup}{\centering}

\begin{table*}[t]
	\tabcolsep=10pt
	\centering
	\caption{Overall comparison results for VGGNet and ResNet-56 on CIFAR-10.}
	\begin{tabular}{c| c| c| c|c| c| c| c}
		\hline

		Models & Methods & Baseline (\%) & Err. Gap (\%) & \#Param. & \#Param. $\downarrow$ & \#FLOPs & \#FLOPs $\downarrow$\\ 
		\hline
		\hline
		
		\multirow{5}*{VGG-16} & CP \cite{He2017} 		   				&  \multirow{4}*{6.01} &   +0.32	&   \multirow{4}*{7.70M}	&   \multirow{4}*{1.92$\times$}	 &   \multirow{4}*{155.80M}	 &   \multirow{4}*{2.00$\times$}\\
							  & ThiNet \cite{Luo2017}	   				&   ~	   &   +0.14	&   ~	&   ~	 &   ~	 &   ~\\
							  & Sliming \cite{Liu2017} 	   				&   ~	   &   +0.19	&   ~	&   ~	 &   ~	 &   ~\\
							  & DCP \cite{zhuang2018discrimination}	    &   ~	   &   -0.17	&   ~	&   ~	 &   ~	 &   ~\\
							  \cline{2-8}
							  & Ours	   								&   6.08   & \textbf{-0.13}	 &   5.50M	&  \textbf{2.26}$\times$ &  140.30M &  \textbf{2.23}$\times$\\
							   \hline
		\multirow{5}*{ResNet-56}  & CP \cite{He2017}	   &   \multirow{4}*{6.20}	   &   +1.00 &   \multirow{5}*{0.42M} &   \multirow{5}*{1.97$\times$}	 &   \multirow{5}*{63.20M}	 &   \multirow{5}*{1.99$\times$} \\
								& ThiNet \cite{Luo2017}   &   ~    &   +0.82	&   ~	&   ~	 &   ~	 &   ~\\
								& WM \cite{howard2017mobilenets}  &   ~	   &   +0.56	&   ~	&   ~	 &   ~	 &   ~\\
								& DCP \cite{zhuang2018discrimination}	   &   ~	   &   +0.31	&   ~	&   ~	 &   ~	 &   ~\\
								\cline{2-4}
								& Ours	   &   6.26	   &   \textbf{-0.24}		&   ~	&   ~	 &   ~	 &   ~\\
		
		\hline
	\end{tabular}
	\label{tab:OverallComparison}
\end{table*}

\begin{table}[t]
	\tabcolsep=10pt
	\centering
	\caption{The comparison results on ResNet-56 with different losses on CIFAR-10.}
	\begin{tabular}{c| c| c}
		\hline
		Methods & Training Err. (\%) & Test Err. (\%)\\ 
		\hline\hline
		
		$\mathcal{L}_r$    &   3.75	   &   9.74\\
		$\mathcal{L}_s$	   &   6.53	   &   10.95\\
		$\mathcal{L}_c$	   &   0.85   &   8.35\\
		$\mathcal{L}_r+\mathcal{L}_s$	   &   1.31	   &   8.27\\
		$\mathcal{L}_r+\mathcal{L}_c$	   &   0.73	   &   8.14\\
		$\mathcal{L}_s+\mathcal{L}_c$	   &   1.03	   &   8.20\\
		$\mathcal{L}_r+\mathcal{L}_s+\mathcal{L}_c$	   &   1.09	   &   8.00\\
		\hline
	\end{tabular}
	\label{tab:Multi-Loss-Compare}
\end{table}

\section{Experiments}
\label{sec:typestyle}

In this section, our approach is evaluated on CIFAR-10 \cite{krizhevsky2009learning} and CIFAR-100 \cite{krizhevsky2009learning}. The proposed method is compared with several state-of-the-art methods including ThiNet \cite{Luo2017}, CP \cite{He2017} and DCP \cite{zhuang2018discrimination}. Our approach is implemented using the PyTorch \cite{paszkepytorch} framework. For the experiments on CIFAR-10 and CIFAR-100, we adopt the same experiments setting as in \cite{zhuang2018discrimination}. Besides, $\alpha$ and $\beta$
(Eq. \ref{eq:FinalLoss}) are set to $0.001$ and $1$ respectively.
\subsection{Overall Comparison Results}
\label{ssec:subhead}

To verify the effectiveness of the proposed approach, we conduct comparison experiments with several state-of-the-art methods on VGGNet and ResNet-56 . Table \ref{tab:OverallComparison} summarizes overall comparison results on CIFAR-10. It can be observed that the accuracy gap between our baseline model and the comparison reference model is very small. For VGGNet, the proposed approach has a higher speedup and compression rate than other methods with the same accuracy drop. For ResNet-56, our approach achieves the best performance compared with previous state-of-the-art methods. With a compression rate of $1.97\times$ and a speedup of $1.99\times$, the proposed method outperforms the DCP by $0.55\%$ in accuracy drop. Moreover, our pruned ResNet-56 even outperforms the baseline model by $0.24\%$ in accuracy.

\subsection{Analysis of the Proposed Approach}
\label{ssec:subhead}

To study effects of different losses, we prune $30\%$ channels of ResNet-56 on CIFAR-10 with different losses without fine-tuning. As shown in Table \ref{tab:Multi-Loss-Compare}, the test error of the pruned model gets an immediate decline when introducing the feature and semantic correlation loss or the classification loss. It's worth noting that the pruned model obtains the best performance with three losses together. This demonstrates multi-loss fusion is an effective strategy for channel pruning and three losses can complement each other in some aspects. 

\renewcommand\arraystretch{1.11}
\begin{table}[t]
	\tabcolsep=10pt
	\centering
	\caption{The comparison results on ResNet-18 and ResNet-34 with different pruning rate on CIFAR-100.}
	\begin{tabular}{c| c| c}
		\hline
		
		Methods & Pruned rate (\%) & Test Err. (\%)\\ 
		\hline\hline
		
		\multirow{3}*{\shortstack{ResNet-18 \\ (Baseline 21.89\%)}}  & 30	   &   24.39\\
		& 50	   &   24.91\\
		& 70	   &   25.15\\
		\hline
		\multirow{3}*{\shortstack{ResNet-34 \\ (Baseline 21.16\%)}}  & 30	   &   21.87\\
		& 50	   &   22.41\\
		& 70	   &   22.89\\			
		\hline
		
	\end{tabular}
	\label{tab:Different_Pruning_Rate}
\end{table}

Going one step further, we verify the convergence of our method. The loss function in Eq. \ref{eq:FinalLoss} is optimized using SGD when pruning a single layer. Fig. \ref{fig:Convergence} shows three losses (Eq. \ref{eq:FinalLoss}) change over epochs for the first twelve layers of VGGNet on CIFAR-100. It can be found that models with different losses can converge to a minor error after $50$ epochs. At Conv5-1 and Conv5-2, curves of three losses do not drop after $10$ epochs. The above results demonstrate the proposed method has a fast rate of convergence. 

To further explore limits of the proposed approach, we prune ResNet-18 and ResNet-34 with different pruning rate on CIFAR-100. Experimental results in Table \ref{tab:Different_Pruning_Rate} show that the pruned model gets an immediate accuracy drop with increase of pruning rate. Besides, ResNet-18 is more sensitive to channel pruning than ResNet-34 due to its fewer parameters.

\section{Conclusion}
\label{sec:print}

In this paper, a multi-loss-aware channel pruning method is presented for deep networks compression. The proposed approach achieves the state-of-the-art performance on benchmark datasets by considering the reconstruction error, the classification loss and the feature and semantic correlation at the same time. As for future works, we will combine existing pruning strategies with other network compression methods to explore more compact models with less accuracy drop.

\section{Acknowledgment}
This work has been supported by the National Key Research and Development Program of China No. 2018YFD0400902 and the National Natural Science
Foundation of China (Grant No. 61673376 and 61573349). 

\bibliographystyle{IEEEbib}

\end{document}